\documentclass[sigconf]{acmart}

\AtBeginDocument{%
  \providecommand\BibTeX{{%
    \normalfont B\kern-0.5em{\scshape i\kern-0.25em b}\kern-0.8em\TeX}}}

\usepackage{amsmath,bm,subfigure,multirow,booktabs,tabularx,threeparttable}

\copyrightyear{2021}
\acmYear{2021}
\setcopyright{acmlicensed}
\acmConference[MM '21]{Proceedings of the 29th ACM International Conference on Multimedia}{October 20--24, 2021}{Virtual Event, China}
\acmBooktitle{Proceedings of the 29th ACM International Conference on Multimedia (MM '21), October 20--24, 2021, Virtual Event, China}
\acmPrice{15.00}
\acmISBN{978-1-4503-8651-7/21/10}
\acmDOI{10.1145/3474085.3475493}

\acmSubmissionID{mfp1780}

\begin{document}
\fancyhead{}

%%
%% The "title" command has an optional parameter,
%% allowing the author to define a "short title" to be used in page headers.
\title{Pairwise Emotional Relationship Recognition in Drama Videos: Dataset and Benchmark}

%%
%% The "author" command and its associated commands are used to define
%% the authors and their affiliations.
%% Of note is the shared affiliation of the first two authors, and the
%% "authornote" and "authornotemark" commands
%% used to denote shared contribution to the research.

% \author{Xun Gao,\quad Yin Zhao$^*$,\quad Jie Zhang,\quad Longjun Cai}
% \makeatletter
% \def\authornotetext#1{
% \if@ACM@anonymous\else
%     \g@addto@macro\@authornotes{
%     \stepcounter{footnote}\footnotetext{#1}}
% \fi}
% \makeatother
% \authornotetext{Corresponding author.}

% \affiliation{
%  \institution{Alibaba Group, Beijing, China}
%  }
% % \email{gaoxun.gx@alibaba-inc.com, yinzhao.zy@alibaba-inc.com, zj209798@alibaba-inc.com, longjun.clj@alibaba-inc.com}
% \email{{gaoxun.gx,  yinzhao.zy, zj209798, longjun.clj}@alibaba-inc.com}

% \def\authors{Xun Gao, Yin Zhao, Jie Zhang, Longjun Cai}

\author{Xun Gao}
% \authornote{Both authors contributed equally to this research.}
\email{gaoxun.gx@alibaba-inc.com}
\orcid{0000-0003-2322-7948}
\affiliation{%
  \institution{Alibaba Group}
  % \streetaddress{P.O. Box 1212}
  \city{Beijing}
  % \state{Ohio}
  \country{China}
  \postcode{10000}
}

\author{Yin Zhao}
\authornote{corresponding author}
\email{yinzhao.zy@alibaba-inc.com}
\affiliation{%
  \institution{Alibaba Group}
  % \streetaddress{P.O. Box 1212}
  \city{Beijing}
  % \state{Ohio}
  \country{China}
  \postcode{10000}
  }

\author{Jie Zhang}
\email{zj209798@alibaba-inc.com}
\orcid{0000-0002-3877-9385}
\affiliation{%
  \institution{Alibaba Group}
  % \streetaddress{P.O. Box 1212}
  \city{Beijing}
  % \state{Ohio}
  \country{China}
  \postcode{10000}
}

\author{Longjun Cai}
\email{longjun.clj@alibaba-inc.com}
\affiliation{%
  \institution{Alibaba Group}
  % \streetaddress{P.O. Box 1212}
  \city{Beijing}
  % \state{Ohio}
  \country{China}
  \postcode{10000}
}

%
% By default, the full list of authors will be used in the page
% headers. Often, this list is too long, and will overlap
% other information printed in the page headers. This command allows
% the author to define a more concise list
% of authors' names for this purpose.
\renewcommand{\shortauthors}{Xun GAO et al.}

%% The abstract is a short summary of the work to be presented in the
%% article.
\begin{abstract}
    
Recognizing the emotional state of people is a basic but challenging task in video understanding.
In this paper, we propose a new task in this field, named Pairwise Emotional Relationship Recognition (PERR).
This task aims to recognize the emotional relationship between the two interactive characters in a given video clip.
It is different from the traditional emotion and social relation recognition task.
Varieties of information, consisting of character appearance, behaviors, facial emotions, dialogues, background music as well as subtitles contribute differently to the final results, which makes the task more challenging but meaningful in developing more advanced multi-modal models.
To facilitate the task, we develop a new dataset called Emotional RelAtionship of inTeractiOn (ERATO) based on dramas and movies.
ERATO is a large-scale multi-modal dataset for PERR task, which has 31,182 video clips, lasting about 203 video hours.
Different from the existing datasets, ERATO contains interaction-centric videos with multi-shots, varied video length, and multiple modalities including visual, audio and text.
As a minor contribution, we propose a baseline model composed of Synchronous Modal-Temporal Attention (SMTA) unit to fuse the multi-modal information for the PERR task.
In contrast to other prevailing attention mechanisms, our proposed SMTA can steadily improve the performance by about 1\%.
We expect the ERATO as well as our proposed SMTA to open up a new way for PERR task in video understanding and further improve the research of multi-modal fusion methodology.

\end{abstract}

%%
%% The code below is generated by the tool at http://dl.acm.org/ccs.cfm.
%% Please copy and paste the code instead of the example below.
%%

%%
%% Keywords. The author(s) should pick words that accurately describe
%% the work being presented. Separate the keywords with commas.

%% A "teaser" image appears between the author and affiliation
%% information and the body of the document, and typically spans the
%% page.

% \begin{teaserfigure}
%  \includegraphics[width=0.95\textwidth]{ERATO_example.pdf}
%  % \vspace{-0.7cm}
%   \caption{Intimate or hostile? Obviously, the woman's facial expression is negative due to the illness. However, their emotional relationship is intimate (positive) because the 
%   man comforts the woman by mildly touching and talking. Best viewed in color.}
%  % \vspace{-0.4cm}
%   \label{fig:ERATO}
% \end{teaserfigure}

%%
%% This command processes the author and affiliation and title
%% information and builds the first part of the formatted document.
\maketitle

\section{Introduction}
Sentiment analysis plays important roles in video understanding.
The typical sentiment analysis involves the tasks such as Facial Expression Recognition (FER) \cite{2005Web,mollahosseini2017affectnet,2013Emotion,2016Facial},
which classifies the facial expression into certain categories;
Group Emotion Recognition (GER) \cite{2018EmotiW,2015The,2016EmotiW,2012Finding},
which predicts the overall emotional state for a group of people;
and even the Audience Affective Responses Recognition, which predicts the audiences' emotional state with Valence-Arousal measures when watching the videos \cite{2017AFEW,baveye2015liris}.

\begin{figure*}[th]
   \includegraphics[width=0.95\textwidth]{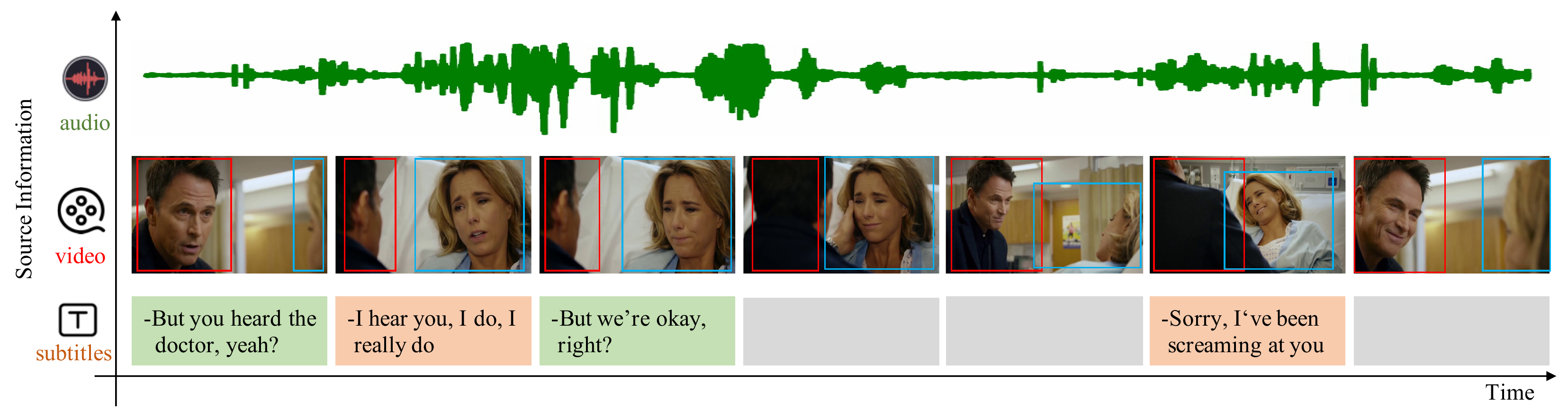}
   % \vspace{-0.4cm}
  \caption{Intimate or hostile? Obviously, the woman's facial expression is negative due to the illness. However, their emotional relationship is intimate (positive) because the 
  man comforts the woman by mildly touching and talking. Best viewed in color.}
  % \vspace{-0.2cm}
  \label{fig:ERATO}
\end{figure*}

Herein we propose a new task named Pairwise Emotional Relationship Recognition (PERR), which is defined as identifying the category of the emotional relationship between two interactive characters in a given video clip.
The emotional relationship can have two scales: one is coarse, including positive, neutral and negative, and the other is fine-grained, including hostile, tense, mild, intimate and neutral.
The given clips contain video, audio, as well as subtitles information.
Fig.\ref{fig:ERATO} gives an illustration of this task. 
A man and a woman are talking to each other before the bed in a sickroom. 
Although the woman seems to be sick, the man is comforting her and the overall emotional relationship is intimate by their interactions such as facial expression, behaviors, dialogues and even the background music.
% This kind of pairwise interaction is widely spread in dramas and movies.
Identifying the emotional relationship for those interactions can advance understanding the video content, especially in the evolvement of the character relations in the storyline, how the story is going on, etc.

PERR task is substantially different from the existing emotion-related or relation-related recognition tasks, such as FER, GER and Social Relation Detection.
FER is generally to predict a single person's facial expression and GER is aiming to predict the overall emotional state of a group of people.
Their labels are the emotion categories or the emotional polarity, not the relationship.
Social relationship generally refers to the relation such as colleagues, couples, etc., while PERR aims at predicting the emotional relationship such as intimate, hostile.
The emotional relationship is relative straightforward compared with those implicit social relationship\cite{2009Understanding,goel2019end,li2017dual,sun2017domain,2015Beyond}.
What's more, PERR is a multi-modal task originating from dramas or movie videos. 
Because of the multi-camera shooting and professional editing techniques,
the video clip generally has multiple shots and overlapping appearance of characters, which make the task more challenging.

To facilitate the PERR task, we develop a new dataset called Emotional RelAtionship of inTeractiOn (ERATO).
ERATO is a large-scale high-quality dataset that has emotional relationship annotations. 
Specifically, it consists of more than 30000 interactive clips extracted from movies or dramas, with annotations of the two people in the interaction, emotional relationship categories as well as subtitles.
Consistent with the PERR task, the ERATO dataset is interaction-centric, with multi-shot, multi-modality and high-quality.
The data collection and annotation process are under strict quality control. 
All the above aspects make ERATO a reliable start point and benchmark for PERR task.
On merit of the ERATO, we propose a simple but effective Synchronous Modal-Temporal Attention (SMTA) unit for PERR task.
SMTA extends LSTM unit using the attention mechanism, including dot-product \cite{wu2019learning} , multi-head attention \cite{vaswani2017attention} or even Transformer encoder\cite{vaswani2017attention} to fuse the multi-modal and temporal context information at the same time.

The contributions of this work are two aspects.
We initialize a new task called Pairwise Emotional Relationship Recognition (PERR) as well as a new dataset called ERATO. PERR is the first task that attempts to predict the emotional relationship for pairwise interaction. 
% ERATO is the first large-scale dataset for PERR and provides a good benchmark for PERR.
% We hope this task and dataset could advance research of emotional relationship recognition and multi-modal learning methods. 
As a start point, we propose a benchmark method using Synchronous Modal-Temporal Attention (SMTA) unit to fuse the multi-modal information in the task.
Compared to the prevailing attention mechanism \cite{wu2019learning,wang2018non,vaswani2017attention}, our SMTA unit can stably improve the performance of PERR on ERATO. 
Code and dataset are available at \url{https://github.com/CTI-VISION/PERR}.

\section{Related Work}
% We present related work in understanding emotion, studying social relationships, and analyzing movies or TV shows for other related tasks. 
% \noindent\textbf{Emotion recognition datasets}. 

\subsection{Emotion Recognition Datasets}
% \textbf{Emotion recognition datasets}. 
Early datasets for FER such as JAFFE \cite{lyons1998coding}, 
CK \cite{tian2001recognizing,2010The}, MMI \cite{2005Web}, and MultiPie \cite{2010Multi} were captured in a lab-controlled environment. 
Thus, the datasets collected in the wild condition, which contains the people’s naturalistic emotion states \cite{2018Deep,2016Facial,2016Tega,dhall2011acted,dhall2011static,mollahosseini2016facial,2013Affectiva,2016EmotioNet,mollahosseini2017affectnet} have attracted much more attention. 
% The Emotion Recognition in the Wild (EmotiW) challenges \cite{2016EmotiW,dhall2017individual,2018EmotiW} proposed several datasets, including facial expression datasets and group emotion datasets.
Specifically, the AFEW \cite{dhall2011acted} includes video clips extracted from movies and SFEW \cite{dhall2011static} is built using static images from the subset video clips of AFEW.
They all use annotations of 7 emotional categories (6 basic emotions plus neutral).
% The group emotion dataset, i.e. HAPPEI/Group Affect Database \cite{dhall2015automatic,2015The} consisted of the websites images with the annotation of the affect conveyed by the group of people. 
The group emotion dataset, i.e. HAPPEI/Group Affect Database \cite{dhall2015automatic,2015The} consisted of the images from the social-networking
websites such as Flickr and Facebook, with the annotation of the affect conveyed by the group of people in the image. 
Recently, datasets for GER in terms of videos are proposed, including CAER \cite{lee2019context}, VGAF \cite{sharma2019automatic} and GroupWalk \cite{2020EmotiCon}.

% All the above datasets have annotations of discrete categories.
However, the human emotional state is complex in the real world. 
The more descriptive measure: Valence and Arousal \cite{russell1980circumplex} are introduced to the emotion datasets.
In ACM-faces \cite{panagakis2015robust}, AffectNet \cite{mollahosseini2017affectnet} and AFEW-VA \cite{2017AFEW}, the values of Valence-Arousal were provided to depict the facial expression in still images. 
The datasets \cite{dhall2012collecting,cowie2000feeltrace,douglas2007humaine,schroder2009demonstration} were created on laboratory or controlled environment using VA value in the level of video clips.
LIRIS-ACCEDE \cite{baveye2015liris} annotated the audience responses for the video clips of movies using Valence and Arousal description.
% This is what the FER and GER cannot do.
% PERR work is fundamentally different from the above literature. 

\subsection{Social Relationship Datasets}
Early studies on predicting social relationships \cite{2009Understanding,goel2019end,li2017dual,sun2017domain,2015Beyond} 
are based on images, including PIPA \cite{2015Beyond}, PISC \cite{li2017dual}, etc.
% Those categories were defined based on the sociological theories.
PIPA contains 16 social relations and PISC has a hierarchy of 3 coarse-level relationships and 6 fine-level relationships. 
Sun et al. \cite{sun2017domain} extended the PIPA dataset and grouped 16 social relationships into 5 categories.
Researchers further constructed video-based dataset for social relation prediction. 
SRIV \cite{lv2018multi} contains about 3000 video clips collected from 69 movies with 8 subjective relations. 
ViSR \cite{2019Social} defines 8 types of social relations derived from the Burgental’s domain-based theory \cite{Bugental2000Acquisition}, containing over 8,000 video clips. 
MovieGraphs \cite{vicol2018moviegraphs} consists of 7637 movie clips annotated with persons, objects, visual or inferred properties, and their interaction including social relationships.
In dramas, the social relationship might be implicit and it might need too much additional information beyond that clip to infer it.
However, the emotional relationship would be more straightforward based on the characters' expression, behavior, dialogue as well as background music.
% We are obsessed with tracking this change and propose the ERATO dataset for PERR.

% \clearpage
\subsection{Methods in Video Understanding}
Herein, we survey the multi-modal approaches for video understanding. 
Most of the current multi-modal methods are model-agnostic approaches \cite{2015A}, which can be divided into early, late and hybrid fusion \cite{2010Multimodal}. 
The difference between the early and late fusion depends on whether the fusion is at the feature level or prediction level.
For instance, Castellano et al. \cite{2008Emotion} used early fusion to combine facial expression, body gestures and 
speech for emotion recognition.
In \cite{2020Learning}, the visual and textual modalities were aggregated after the prediction of each modal. 
While those model-agnostic approaches are relatively easy to implement, they are generally not effective in mining the multi-modal data.

Recently, the model-based approaches explicitly designed the architecture based on different properties of the tasks. 
The Attention mechanism is the main choice to fuse the multi-modal information.
The Non-local \cite{wang2018non} operator can capture the long-range dependencies and has been verified in the video action recognition. 
Alc{\'a}zar et al. \cite{alcazar2020active} applied the non-local operation in the task of active speaker detection.
The non-local operation ensembled the audio-visual context while the subsequent LSTM fused the temporal context information. 
% Transformer or Bert \cite{vaswani2017attention,devlin2018bert}, which are the popular architectures in the natural language process, are based on self-attention mechanism.
Transformer or Bert \cite{vaswani2017attention,devlin2018bert} also have excellent performance on multi-modal tasks in video understanding \cite{khan2021transformers}, 
such as multi-modal representation \cite{lu2019vilbert,lee2020parameter,sun2019videobert} and video captioning \cite{zhou2018end}. 
In VideoBERT \cite{sun2019videobert} and ViLBERT \cite{lu2019vilbert}, a modality-specific transformer encoded long-term dependencies for individual modality, then a multi-modal transformer was followed to exchange the information across visual-linguistic modalities. 

Since the proposed PERR task is a video-based multi-modal problem, the fusion methods should consider two dimensions, one is for different modalities, and the other is for time dependency.
In this aspect, \cite{alcazar2020active} dealt with modality fusion first and then considered time dependency, while \cite{sun2019videobert,lu2019vilbert} took the reverse order.
To provide a baseline, we proposed a method that uses Synchronous Modal-Temporal Attention unit, considering the time dependency when fusing the multi-modal information.

\section{PERR and ERATO Dataset}
\subsection{Problem Description}\label{sec:problem_def}
We expect to recognize the emotional relationship between the two interactive characters in the video of dramas or movies.
There is no category definition in the emotional relationship. 
Motivated by the related work of GER and the existing human emotion researches Emotional Intimacy \cite{Alan1972} and Emotional Climate \cite{de1992emotional}, 
we proposed two emotional relationship category definitions.
The first one is neutral, positive and negative, coarsely describing the emotional relationship.
The second definition is fine-grained, including mild/intimate/tense/hostile considering that they are commonly listed in above literatures.
Mild and intimate, tense and hostile are fined-grained labels for positive and negative in the coarse category definition shown in Table \ref{tab:category_desc}.

% In sentiment analysis field, FER task generally has 6 to 8 categories including neutral, happy, sad, surprise, fear, anger, disgust. 
% Three categories, i.e. neutral, positive, negative are devoted to describe the group emotion. 
% In view of the fact that pariwise interaction is close to GER task, we also use neutral, positive and negative to coarsely describe the emotional relationship.
% Positive means the emotional relationship between the two characters is friendly, harmonious, imitate; while negative means the two characters' 
% emotional relationship is tense, hostile, etc.
% % In addition, 
% After referring to 
% we further refine the positive and negative into different intensities.
% For the positive, we divide it into mild and intimate while for the negative, we divide it into tense and hostile as shown i

% Recognizing the pairwise interaction and its emotional relationship is a classification task involving many stages.
For untrimmed video like dramas and movies, the audiences can easily identify who are the main interactive characters in a certain period and further recognize whether they are intimate or hostile.
However, it is much more challenging for machines since it involves person detection, main character identification, person tracking, interaction detection, multi-modal feature extraction, multi-modal information fusion, etc.
To pay more attention on the multi-modal fusion, we define the PERR task as follows:
For a given interaction-centric video clip, we have known the main interaction characters and their textual dialogues.
What we want to do is to classify the emotional relationship between the specified two main characters in this clip into the defined emotion relationship class.

\begin{table}[tbp]
\small
\caption{The category definition in PERR.}
% \vspace{-0.4cm}
\centering
\begin{tabularx}{\linewidth}{cXX}
\toprule
        \multicolumn{2}{c}{\textbf{Category}}                                                  & \multicolumn{1}{c}{\textbf{Description}} \\ 
\midrule

\multicolumn{1}{c|}{\multirow{3}{*}{Negative}}  & \multicolumn{1}{c|}{\multirow{2}{*}{Hostile}} & 
                                                  Antagonistic relationship, usually in fierce quarrel or even physical conflict \\  \cline{2-3} 
\multicolumn{1}{c|}{}                          & \multicolumn{1}{c|}{\multirow{1}{*}{Tense}} & Disagree with each other or under argument    \\ \hline

\multicolumn{1}{c|}{\multirow{3}{*}{Positive}} & \multicolumn{1}{c|}{\multirow{1}{*}{Mild}}  & Heart-warming, pleasant dialogue with smile \\ \cline{2-3} 
\multicolumn{1}{c|}{} &  \multicolumn{1}{c|}{\multirow{2}{*}{Intimate}}                        & 
                                                  Close interactions with an affectionate or loving manner, including intimate physical contact \\ \hline
  
\multicolumn{2}{c|}{\multirow{2}{*}{Neutral}}                                                  &Neutrality, normal dialogue without emotional disposition \\ 

\bottomrule
\end{tabularx}
% \vspace{-0.4cm}
\label{tab:category_desc}
\end{table}

% \begin{figure*}[htbp]
% \addtocounter{figure}{-1} %%%%%%%%%note1
\begin{figure*}[htbp]

\centering
\subfigure[the distribution of clip duration]{
\begin{minipage}[t]{0.25\linewidth}
\centering
\includegraphics[width=\linewidth]{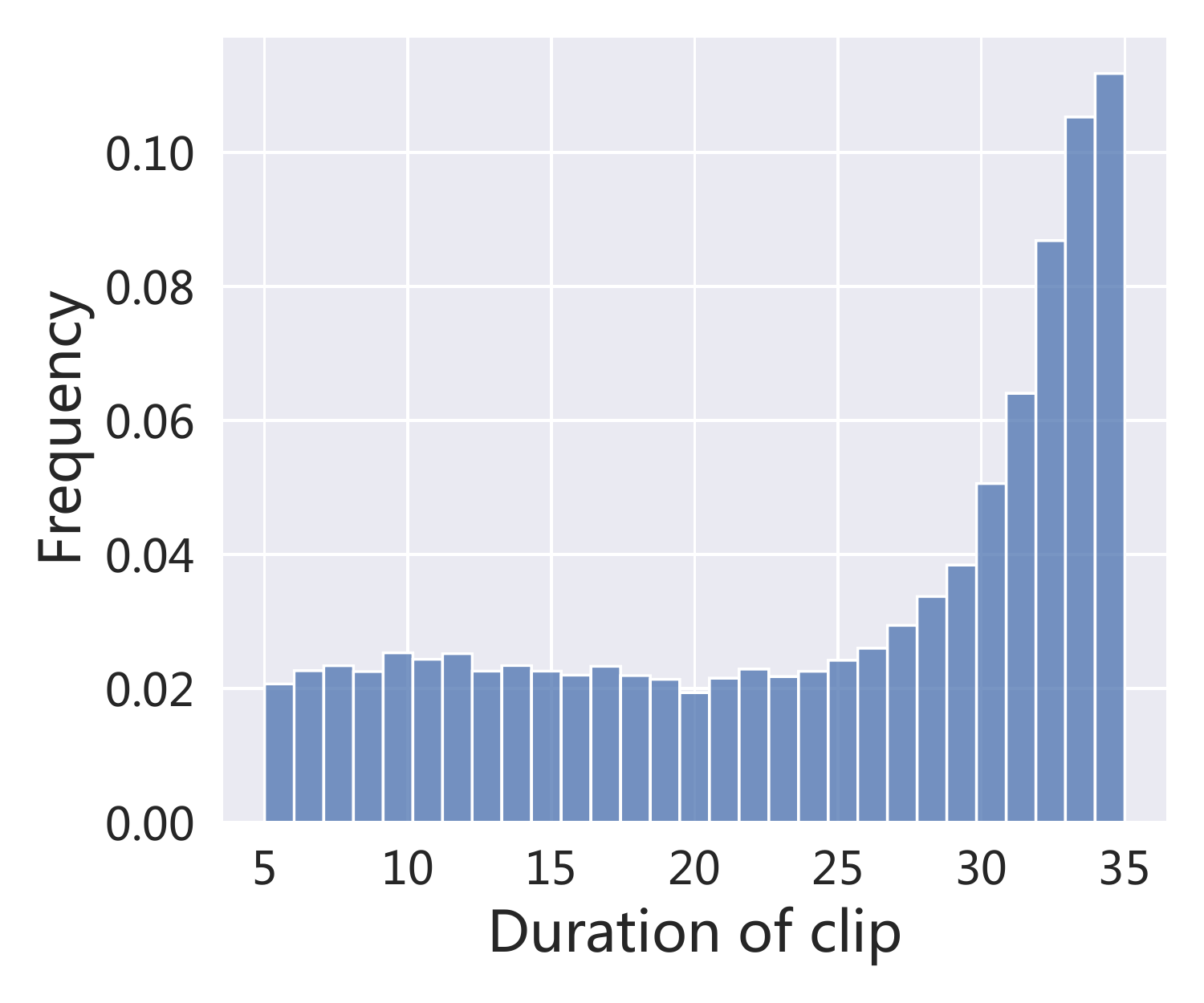}
% \caption{fig1}
\end{minipage}%
}%
\subfigure[the distribution of the shot number]{
\begin{minipage}[t]{0.25\linewidth}
\centering
\includegraphics[width=\linewidth]{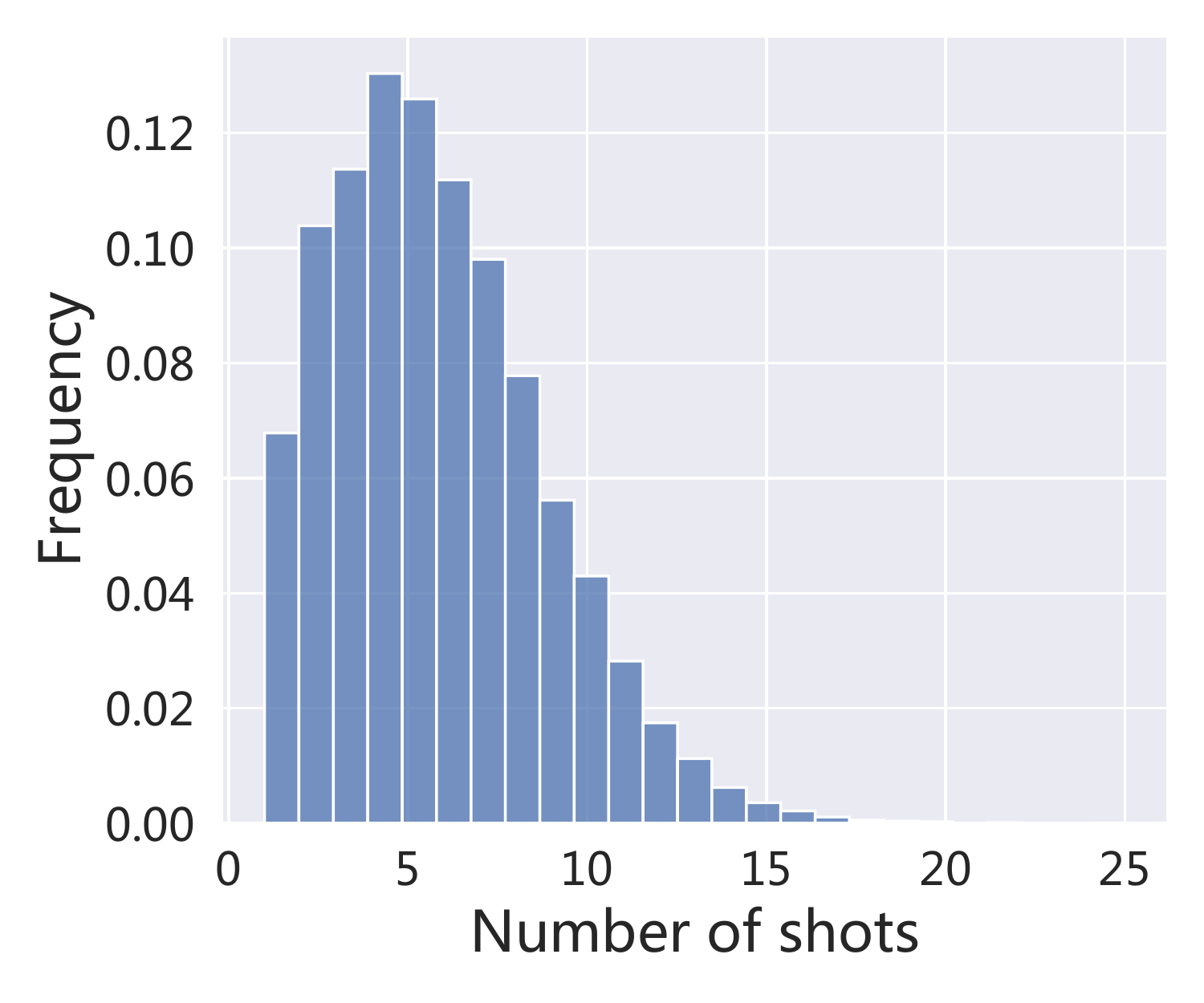}
%\caption{fig2}
\end{minipage}%
}%
\subfigure[the distribution of appearance ratio]{
\begin{minipage}[t]{0.25\linewidth}
\centering
\includegraphics[width=\linewidth]{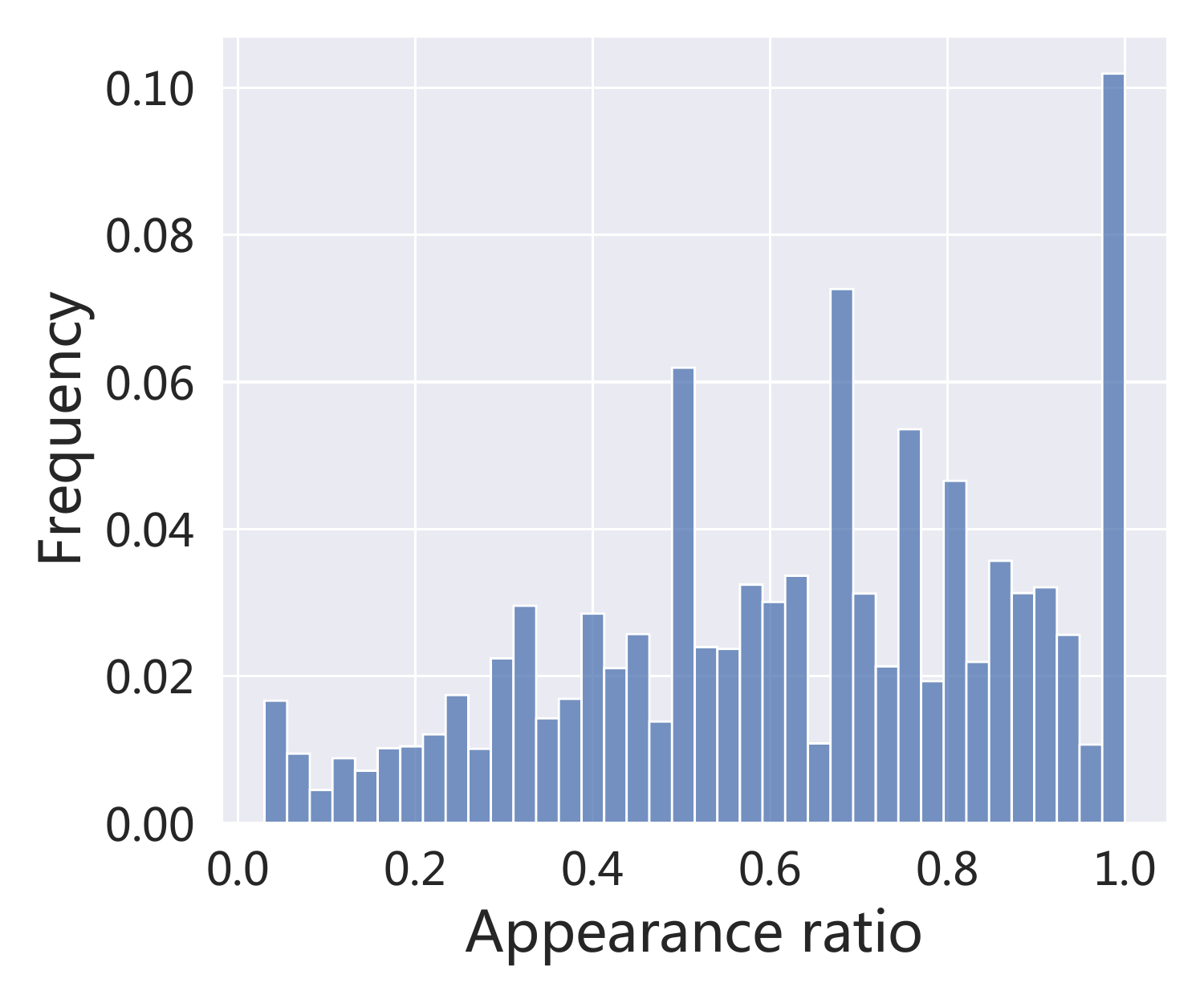}
%\caption{fig2}
\end{minipage}
}%
\subfigure[the distribution of the subtitle ratio]{
\begin{minipage}[t]{0.25\linewidth}
\centering
\includegraphics[width=\linewidth]{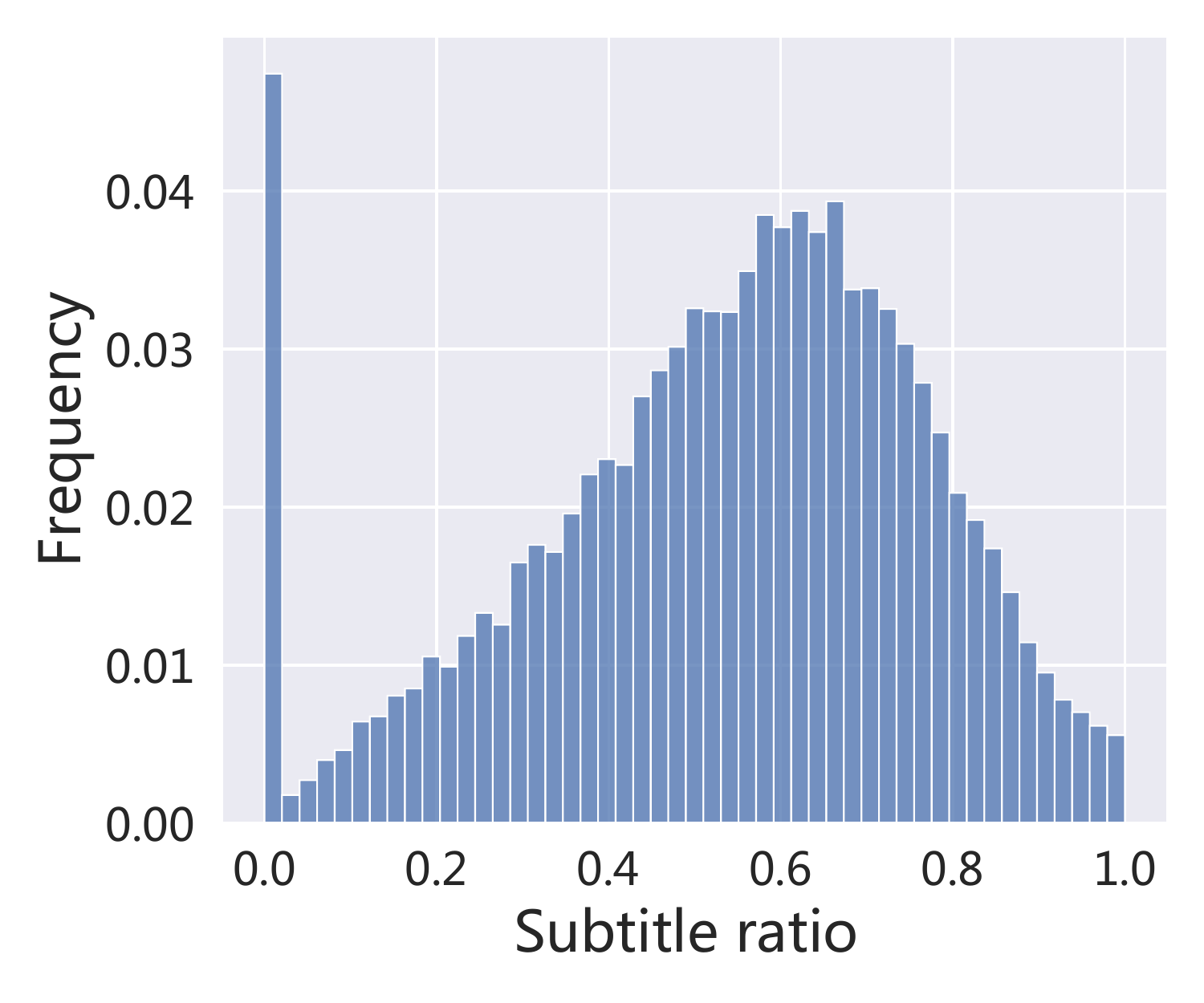}
%\caption{fig2}
\end{minipage}
}%
\centering
% \vspace{-0.2cm}
\caption{Statistics of ERATO dataset.}
% \Description{We show that: (a) the distribution of instance length. 
%         (b) the distribution of the shot number that clip contains. (c) the histogram of appearance ratio, 
%         which defined as the ratio of appearance time between paired interactive characters. 
%         (d) subtitle ratio which measures the time-length of text in clip.
%         }
% \vspace{-0.2cm}
\label{tab:dataset_statistic}
\end{figure*}

\subsection{Dataset Construction}\label{sec:data_construction}
% We created the ERATO dataset to facilitate the proposed PERR task.
% In this section, we will describe the data collection and annotation rules.
% The statistics and properties of ERATO are also presented.

\noindent\textbf{Data Preparation.}
To extract the pairwise interactive video clips, we select dramas and movies that have different genres to ensure data diversity.
In total, we have 487 dramas belonging to 8 genres, with 1161 episodes. 
% Each episode is a single video with multiple shots.
We first divide the video into multiple shots using PySceneDetect in \cite{pyscenedetect}.
Then by combining continuous shots, each episode can be roughly divided into story segments like \cite{tapaswi2014storygraphs}.
Note that the resulting story segments generally have a long time duration with many pieces of pairwise interactions.
We manually locate the pairwise interactions from those story segments such that each video clip corresponds to one pairwise interaction.

After the pairwise interaction retrievement, we apply several basic models to extract the basic elements to improve the final annotation efficiency.
Specifically, to help the annotators to identify the two interaction characters, we first detect the faces by MTCNN \cite{zhang2016joint} and track the faces according to the discriminative face features extracted by ArcFace \cite{deng2019arcface}. 
Then we drop the clips that include only one or more than 6 characters.
The top 3 faces (if it has more than 2 faces) with the largest occurrence number are selected as the candidates for the main interactive characters in this clip.
Those 3 faces would be all displayed for the annotators to manually select the main interactive character pair.
Second, we use CTPN \cite{tian2016detecting} and CRNN \cite{fu2017crnn} to detect and recognize the text in the video subtitles. 
The duplicate subtitles in the adjacent frames are removed based on Levenshtein Distance \cite{levenshtein1966binary}.
Those textual dialogue information would also be provided to the annotators to fix the potential errors of the detection and recognition.
In this way, we produce more than 40000 video clips, with the candidate main interactive characters and the candidate subtitles for the following annotations.

\noindent\textbf{Annotation.} 
To guarantee the quality of the dataset, we adopt a series of quality control mechanisms during the annotation. The annotation is divided into two periods: the training period and annotating period. 
Six paid annotators are involved in the process. 
In the training period, we first provide annotation guidance and demos for them.
Then the annotators will annotate the training video clips and cross-check each other's annotations until reaching a consensus.
% After the training period, the formal annotation period starts.
In annotating period, we adopt a multi-stage strategy, which is relatively time-consuming 
but can ensure the quality of the dataset. 
First, annotators need to watch the videos and fine-tune the detected subtitles.
This stage, on the one hand, can correct the potential errors by the text detection and recognition models, on the other hand, will give the annotators an overall impression of the content of the video clip. 
If the pattern of the interaction is complicated (no salient interactive characters, etc.), the annotators are allowed to discard this clip.
Second, the annotators need to select the two main interactive characters from the detected top 3 candidate face images by watching the video again.
Third, according to the interaction between the paired characters, such as facial expression, pose, dialogue information, background music, etc., annotators are asked to choose one of the detailed 5 tags that best reflects the emotional relationship.
During the annotation period, a weekly cross-check would be carried out to ensure the consensus.
The percent agreement \cite{Stephanie2016} was 88.3\% and 82.6\% for the coarse and the fine-grained categories respectively in the early stage, and then gradually increased to 94.8\% and 90.1\% respectively.

\subsection{Dataset Statistics}\label{sec:dataset_stats}
% In the following, we present some basic statistics of the ERATO.

\noindent\textbf{Duration Statistic.} 
ERATO dataset consists of 31,182 annotated video clips, which lasts 203 hours in total.
They are extracted from 487 dramas, involving 1161 episodes. 
Fig. ~\ref{tab:dataset_statistic} (a) depicts the histogram of the clip durations. 
The length of clip varies from 5 to 35 seconds with an average of 23.5 seconds. 
Most of the video clips last for 27 to 35 seconds, which is believed to have enough information to reflect the emotional relationship.

\noindent\textbf{Category Distribution.} 
Table~\ref{tab:cls_statistic} shows the distribution of the emotional relationship categories in ERATO dataset. 
Overall, the neutral category contains more than half of the samples, which means there exists unbalancing category phenomenon in ERATO.
Furthermore, the positive and negative relationships have close ratios, around 19-20\% respectively.
The stronger categories such as hostile and intimate are less than the intermediate categories, i.e. mild and tense.
% As a whole, the resulting distribution is consistent with the distribution in real-world life.
Note that the unbalanced category phenomenon mentioned above is unavoidable, which always exists in real-world life.
To reflect this, we propose to use Micro-F1 and Macro-F1 as the model evaluation metrics, which will be described in Sec.\ref{sec:evaluation}.

\noindent\textbf{Distribution of shot number.} 
Multi-Shot is the unique characteristic of TV shows and movies caused by multi-camera shooting and professional editing techniques. 
From Fig. ~\ref{tab:dataset_statistic} (b), 
it can be observed that the shot number of most video clips ranges from 1 to 9, and the maximum can reach 25. 
The frequent shot changes in the video clip will bring more challenges for the task since the characters and scene will be changed in time.
More advanced methods modeling this interactive phenomenon would be needed.

\noindent\textbf{Appearance ratio of the paired characters.} 
The Appearance ratio is the percentage of the occurrence time (in second) between the paired interactive characters, with the smaller value as the numerator and the other as the denominator. 
Fig. ~\ref{tab:dataset_statistic} (c) shows the histogram of the appearance ratios.
We can see that in about 10\% of the clips, the main interactive characters have equivalent appearance times.
For the left, the appearance time between them is unequal, most varied from 0.2 to 0.9. 
The inequivalence of the two characters might require the model to consider more explicitly 
the different weights or contributions of the two characters in the emotional relationship.

\begin{table}[tbp]
\small
\caption{Sample statistics per category.}
% \vspace{-0.4cm}
\centering
\begin{tabular}{c|c|c|c}
\toprule
\multicolumn{2}{c}{Category}             & \multicolumn{1}{c}{Number}    & Ratio         \\ 
\midrule
\multirow{2}{*}{Negative}    & Tense     & 6088      &  19.52\%      \\ 
                             & Hostile   & 1421      &  4.56\%       \\ \hline
\multirow{2}{*}{Positive}    & Mild      & 3969      &  12.73\%      \\
                             & Intimate  & 1211      &  3.88\%       \\ \hline
\multicolumn{2}{c|}{Neutral}             & 18493     &  59.31\%      \\ 
\bottomrule
\end{tabular}
% }
% \vspace{-0.2cm}
\label{tab:cls_statistic}
\end{table}

\begin{figure*}[htbp]
\centering
\includegraphics[width=0.95\linewidth]{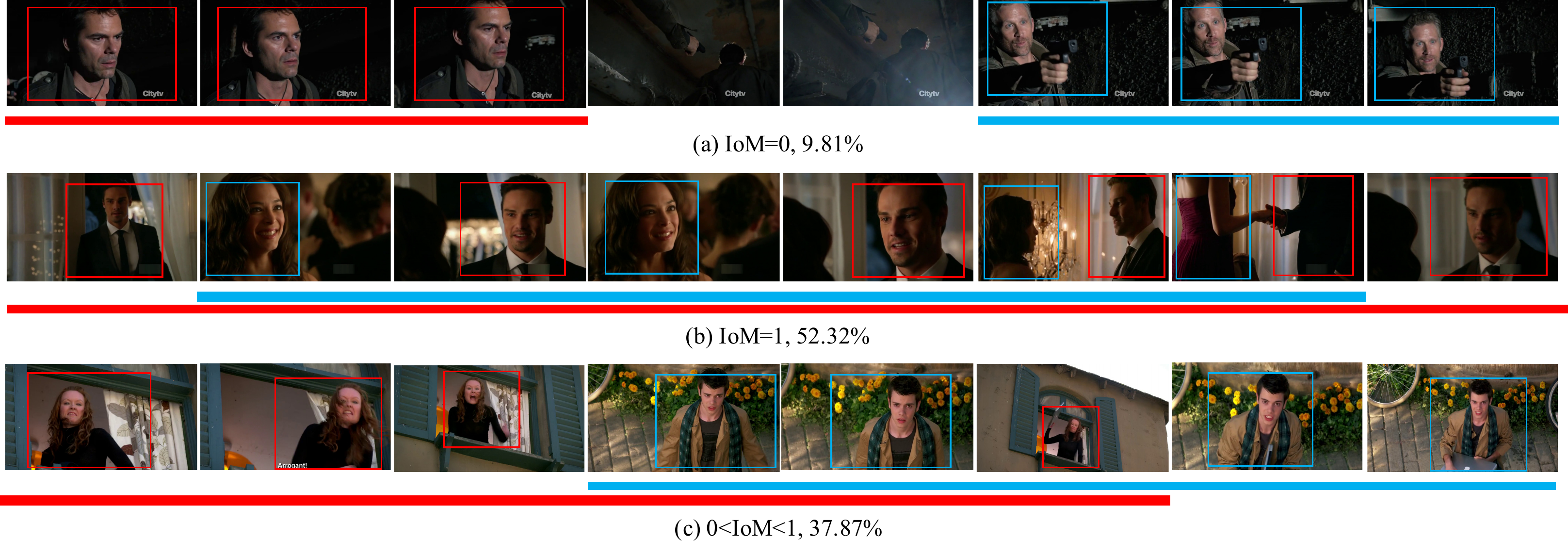}
% \includegraphics[width=\linewidth]{inter_mode_example.pdf}
% \vspace{-0.6cm}
\caption{Examples for different interactive modes in multi-shot video clip. The red and blue line indicate the time span of the two characters, 
and the sub-caption consists of interaction mode and its percentage in ERATO.}
% \Description{}
% \vspace{-0.2cm}
\label{tab:inter_mode_example}
\end{figure*}

% Generally, the smaller the appearance ratio is, the more difficult to emotional relationship recognition, which can promote the research of context information.

% \makeatletter\def\@captype{figure}\makeatother
% \begin{minipage}{.4\linewidth}

% % \includegraphics[width=\linewidth]{interactive_mode.pdf}
% \caption{The percentage of interactive mode.}
% % \Description{}
% \label{tab:interactive_mode}
% \end{minipage}
% \hspace{.1\linewidth}
% \makeatletter\def\@captype{table}\makeatother
% \begin{minipage}{.4\linewidth}
% \centerline{\begin{tabular}{c|c|c}
% \toprule
% \multicolumn{2}{c}{Category}        & Number \\ 
% \midrule
% \multirow{2}{*}{negative}  & tension  & 6088   \\ 
%                           & hostile  & 1421   \\ \hline
% \multirow{2}{*}{positive} & wramth   & 3969   \\
%                           & intimacy & 1211   \\ \hline
% \multicolumn{2}{c|}{neutral}         & 18493  \\ 
% \bottomrule
% \end{tabular}}
% \caption{Sample statistics per category.}
% \label{tab:cls_statistic}
% \end{minipage}
% ~\\
% ~\\

\noindent\textbf{Distribution of Interaction Mode.} 
% The interaction mode reflects the way that the paired 
% characters appear in the video clip, either at the same time, or crossover, or in sequence. 
We also categorize the interaction into different modes by Intersection over Minimum ($IoM$).
% Based on the appearance time of each character, we could have the time span of each character.
$IoM$ calculates the intersection of the two characters' time span over the minimum span of the two characters.
According to the value of $IoM$, we divide the interaction mode into three categories, $IoM$=0, 0<$IoM$<1 and $IoM$=1.
As Fig. ~\ref{tab:inter_mode_example} shows, the sample of $IoM$=0 accounts for 9.81\%, which means that the time span of the pairwise characters does not overlap. 
The proportion of 0<$IoM$<1 is 37.87\%, representing that there is a partial interaction between the two paired characters. 
The situation with the $IoM$=1 is that the appearance time between one of the two characters is inclusive with respect to the other.
The ratio is about 52.32\%.

\noindent\textbf{Distribution of subtitle text.}
Dialogue is important information in ERATO. 
Fig. ~\ref{tab:dataset_statistic} (d) shows the histogram of the ratios of the dialogue duration over the length for each video.
This ratio can be calculated by counting the duration of the subtitles based on the observation that the dialogues and subtitles are synchronized.
We can see that in ERATO, about 4.7\% of the video clips lack the dialogue information, which means the interaction only happens through visual modality: human facial expression and behaviors.
Besides that, the most frequent ratio is from 50\% to 70\%. 
The imbalanced duration of text and visual modalities in ERATO might encourage the development of multi-modal attention or alignment method.
% The misalignment between text and vision is caused by video production, which encourages the development of multimodal attention or alignment.

\begin{figure*}[htbp]
\includegraphics[width=0.95\linewidth]{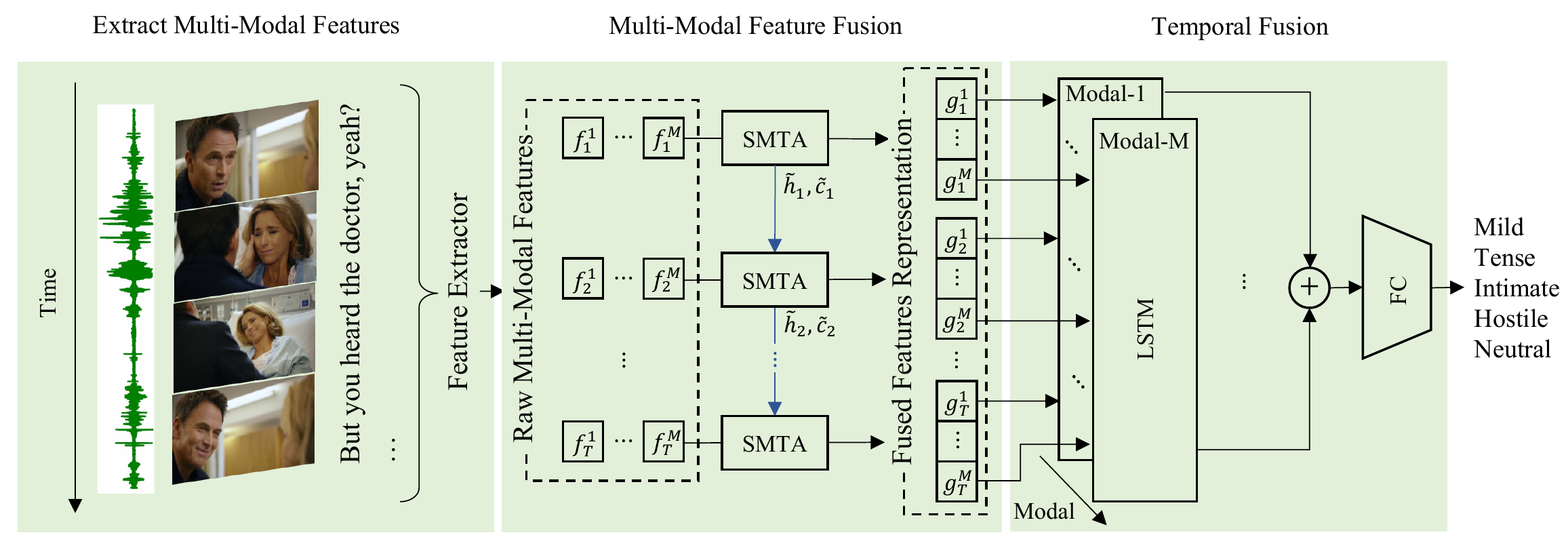}

\centering
% \vspace{-0.3cm}
\caption{Overall framework of our proposed method. The structure of STMA is shown in Fig. ~\ref{tab:general_SMTA}.
}
% \vspace{-0.3cm}
\label{tab:proposed_model}
\end{figure*}

\noindent\textbf{Training and test splitting.} 
We randomly sample 20\% of each category as test set.
In this way the class-wise distribution of the training and testing set is consistent.

\subsection{Dataset Properties}
% ERATO dataset has several distinguished properties.

\noindent\textbf{Interaction-centric.} 
Interaction is the unique and fundamental difference between PERR and other emotion recognition tasks. 
Unlike the tasks such as FER and GER, PERR focuses on interacting characters, not the emotional states of the single character or the group of people.
As shown in Fig.\ref{tab:inter_mode_example}, the samples in ERATO have interactive patterns with multiple shots.
One character first appears in one shot and the other character will appear in different shot.
Sometimes both of them appear at the same time.
This requires new methods to deal with these interactive features.

\noindent\textbf{Multi-modality.} 
ERATO is a video-based dataset thus it is the inherently multi-modal dataset.
ERATO contains visual, audio as well as textual information with start and end timestamps, which are detected by models and refined by annotators.
Most of the existing datasets for emotion recognition do not have textual information. 
Besides facilitating the prediction of the interactive emotional relationship, ERATO can push forward the research on vision-language multi-modal fusion methods.

\noindent\textbf{High-quality and Diversity.} 
Besides the comprehensive quality control mechanism depicted in Sec.\ref{sec:data_construction}, 
the video clips have  high-resolution (e.g. 720P and 1080P) and range over
large genres, including "crime", "war”, "action", "romance", "fantasy", "life", "comedy", and "science fiction".
% \subsection{Evaluation Metrics}

\subsection{Evaluation}
\label{sec:evaluation}
To evaluate the performance of PERR, we employ 2 variants of F1-scores, i.e. Macro-F1 and Micro-F1, considering the imbalanced categories.
Macro-F1 is the arithmetic mean of the per-class F1-scores, which gives equal weights to each class. 
Micro-F1 is the harmonic mean of precision and recall calculated over all the samples without considering the class weights.
By comparing the two metrics, we would evaluate the model performances on the overall test set as well as on each class, especially the minor class.

% Moreover, to illustrate the importance of multi-modality to PERR, we further resort to F1 with different ratio of subtitles. 
% To be concrete, we divide all the instances into 2 groups according to the subtitle ratio and report $\mathrm{F1}_{\mathrm{less}}$ 
% (the ratio less than 0.3) and $\mathrm{F1}_{\mathrm{more}}$ (the rest), respectively.

% \subsection{Potential Applications}
% Based on the current multimodal annotation, the high-quality data of ERATO provides advantages 
% for various applications, besides coarse and fine PERR, including but not limited to: 
% (1) who is interacting? With the annotation of main pairwise interactive characters, ERATO can as 
% a benchmark for localizing the pair of characters that performed interaction. (2) feature representation 
% based on multi-modality. ERATO provides multi-modal sources, including visual video, subtitle text, and 
% audio, which can promote multimodal representation learning. (3) applications for trailer generation 
% and video summary. According to the interactive emotional relationship, it can present the highlights 
% to tease the audience, which is either a pleasant sweet clip, or an exciting quarrel and fight between 
% pairwise characters.

% In this paper, we focus on the emotional relationship of interaction, given the main interactive characters. 
% Specific methods are described below.

\section{Approach}\label{sec:approach}
% On the ERATO dataset, we propose a baseline approach, which opens up the way for further research and reveals the challenges of ERR.
In this section, we present a baseline method for recognizing the emotional relationship based on the ERATO.
The proposed method utilizes a Synchronous Modal-Temporal Attention unit to carry out multi-modal fusion.

\subsection{Overall structures}
Following the definition of PERR in Sec.\ref{sec:problem_def}, we utilize visual, audio as well as textual features to predict the emotional relationships.
Formally, for each video clip with class $y$, denote $f_t^m \in \mathbb R^d$ as the feature of modal $m$ at time $t$ with dimension $d$, $t \in \left \{1, 2, \cdots, T \right \} $ and $m \in \left \{ 1, 2, \cdots, M \right \} $, where $M$ is the number of modalities.
The modalities we use are the character's facial expression, body gesture, subtitle textual features, audio features of the videos.
We extract total of $T$ frames for each video clip.
The details for extracting multi-modal features will be described in Sec. \ref{sec:multi-modal-features}.
Note that the features $f_t^m$ are not available for all the time $t$ as we analyzed in Sec.\ref{sec:dataset_stats}, we use zero-padding for the missing features.

Having the features $f_t^m$, we propose a framework that consists of two stages, the first one is multi-modal feature fusion, the other is time-dependency fusion.
As shown in Fig.\ref{tab:proposed_model}, the multi-modal feature fusion stage takes as input the raw multi-modal features and outputs the fused features representation at different time steps for each modality.
The fused representation considers both the information of the history and different modalities.
The second stage takes as input the modal-wise fused representation at different timesteps and outputs the final prediction.

\subsection{Synchronous Modal-Temporal Attention}
\begin{figure}[]
  \centering
  \includegraphics[width=\linewidth]{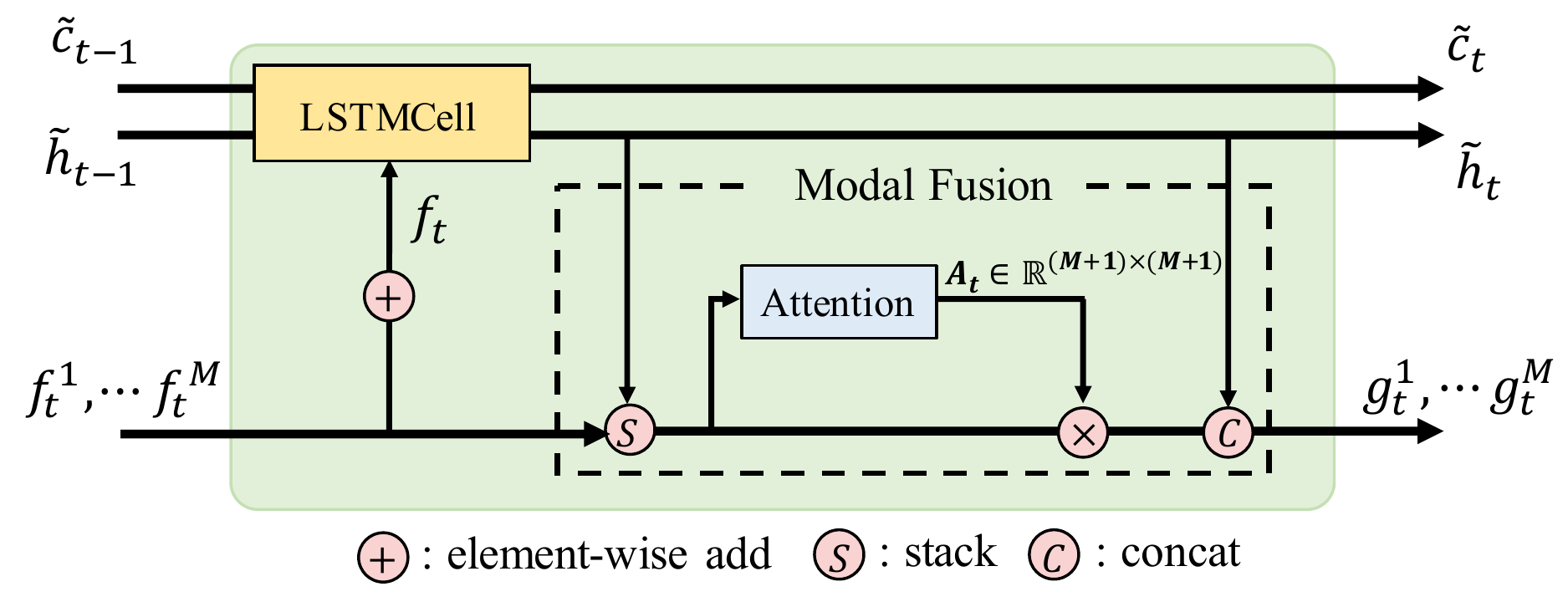}
  % \vspace{-0.6cm}
  \caption{The general structure of Synchronous Modal-Temporal Attention (SMTA) unit.}
  % \Description{}
  % \vspace{-0.4cm}
  \label{tab:general_SMTA}
\end{figure}

In the multi-modal fusion stage, we propose a Synchronous Modal-Temporal Attention (SMTA) unit that can fuse the raw multi-modal features considering the time dependency.
SMTA is a stackable unit extending the LSTM cell.
Fig.\ref{tab:general_SMTA} gives the detailed structure of SMTA.
In detail, SMTA at timestep $t$ takes as input raw features $f_t^m, m=1,2,\cdots, M$, hidden state $\tilde h_{t-1}$ and cell state $\tilde c_{t-1}$ one step before.
Besides the modal-fused features $g_t^m, m=1,2,\cdots, M$, it also outputs the updated hidden state $\tilde h_t$ and $\tilde c_t$, which can be the input of SMTA in the next step.
SMTA consists of two major components, LSTM cell and modal fusion component.
LSTM takes the hidden state $\tilde h_{t-1}$, cell state $\tilde c_{t-1}$ and the sum of raw features of each modality at time $t$, i.e. $f_t = \sum_{m=1}^{M}{f_t^m}$ as the input,
outputing the hidden state and cell state in the next step:
\begin{equation}
\label{tab:SSTA}
  \begin{split}
    \tilde{h}_t, \tilde{c}_t & = \mathrm{LSTMCell} \left ( \tilde h_{t-1}, \tilde c_{t-1}, f_t \right ), \\
    % \tilde{f_t} &= g \left (\tilde{h_t}, f_t \right ) 
  \end{split}
\end{equation}
The modal fusion component takes as input the raw modal features $f_t^m, m=1,2,\cdots, M$ and the updated hidden state of LSTM $\tilde h_t$.
 % i.e. $[f_t^1, f_t^2, \cdots, f_t^M, \tilde h_t]$.
It outputs modal-fused feature representations $g_t^m, m=1,2,\cdots, M$.
We expect the module to consider both the modalities and time-dependency.
To achieve this, we first stack all the input together, i.e. $Z_t = [f_t^1, f_t^2, \cdots, f_t^M, \tilde h_t], Z_t \in \mathbb R^{(M+1)\times d}$.
Then the attention matrix $A_t \in \mathbb R^{(M+1)\times(M+1)}$ is designed to represent the attention weight considering all the modalities as well as the temporal context $\tilde h_t$.
$A_t$ can be implemented in many ways, including the Dot-product attention, Multi-head attention as well as the Transformer encoder. The implementation details are in the appendix and the results are in Table \ref{tab:performance}.
Then we have the intermediate fused features $\tilde Z_t = A_t \cdot Z_t$.
By taking the first $M$ vectors of $\tilde Z_t$ corresponding to $f_t^m, m=1,2,\cdots, M$ and concating with $\tilde h_t$, we have the final fused features, i.e.
$g_t^m = [\tilde h_t, \tilde Z_t^m], m=1,2,\cdots, M$.
Note that direct concatenating $\tilde h_t$ with the intermediate fused features $\tilde Z_t$ takes 
the idea of residual learning in ResNet \cite{2016Deep}.

\subsection{Temporal Fusion}
Having obtained the final fused features $g_t^m, m=1,2,\cdots, M$, we apply LSTM to further fuse the temporal information for each fused modality.
\begin{equation}
\label{tab:lstmcell_}
h_t^m, c_t^m = \mathrm{LSTMCell} ( h_{t-1}^m, c_{t-1}^m, g_t^m  ), m=1,2,\cdots,M
\end{equation}
where $h_t^m$ is the hidden state,
The final prediction of this clip is 
\begin{equation}
\hat y = softmax(FC(\sum_m {h_T^m}))
\end{equation}
where $h_T^m$ is the hidden state of modality $m$ at final timestep $T$, and $FC$ indicates fully connected layer.
In addition, we also calculate the prediction only using one modality, i.e.
\begin{equation}
\hat y^m=softmax(FC(h_T^m))
\end{equation}

The overall objectives of the method can be written as follows:
\begin{equation}
\label{tab:optimize}
 \min \ \ \mathcal{L}(\hat{y}, y) + \sum\nolimits_{m=1}^{M}{\mathcal{L}(\hat{y}^m, y)}
\end{equation}
where $\mathcal{L} \left (\cdot \right )$ indicates the cross-entropy loss.

\section{Experiments \label{Experiments}}

\subsection{Implementation Details} \label{sec:multi-modal-features}

\noindent\textbf{Multi-modal Features}
We extract visual, audio as well as textual features for the PERR task.
The visual features include background features, person-wise features.
The person-wise features consist of facial expression features, posture features as well as the relative position to the other person.
Combining the textual feature and audio feature, there are 5 modal features as the input.
The detailed feature extractions are in the appendix.

\begin{table}[]
\caption{Performance of different methods on ERATO.}
\label{tab:performance}
% \vspace{-0.4cm}
 % \makebox[\linewidth][c]{
\setlength{\tabcolsep}{1mm}{
% \resizebox{\linewidth}{21mm}{
\begin{tabular}{cccccc}
\toprule
\multirow{2}{*}{Method}                           & \multicolumn{2}{c}{5 Category}              & \multicolumn{2}{c}{3 Category}         \\ 
% \toprule
% \hline
                                                  & Micro-F1          & \multicolumn{1}{c}{Macro-F1} & Micro-F1     & Macro-F1   \\ 
% \toprule
\midrule

\multicolumn{1}{c|}{LSTM}                            & 63.80            & \multicolumn{1}{c|}{44.43}    & 64.98        & 56.23      \\ \hline

\multicolumn{1}{l|}{Dot-product \cite{wu2019learning}}      & 64.35            & \multicolumn{1}{c|}{45.80}    & 66.55        & 58.89      \\
\multicolumn{1}{l|}{Non-local \cite{wang2018non}}        & 63.72            & \multicolumn{1}{c|}{47.16}    & 67.49        & 59.74      \\
\multicolumn{1}{l|}{Multi-head \cite{vaswani2017attention}}       & 64.11            & \multicolumn{1}{c|}{47.15}    & 67.30        & 60.20      \\
\multicolumn{1}{l|}{Transformer \cite{vaswani2017attention}}      & 64.82            & \multicolumn{1}{c|}{48.90}    & 68.59        & 58.98      \\ \hline

\multicolumn{1}{l|}{SMTA$_\text{{Dot-product}}$}      & 65.45            & \multicolumn{1}{c|}{47.92}    & 69.18        & 60.20      \\
\multicolumn{1}{l|}{SMTA$_\text{{Non-local}}$}        & 65.33            & \multicolumn{1}{c|}{48.00}    & 69.66        & 60.80      \\
\multicolumn{1}{l|}{SMTA$_\text{{Multi-head}}$}       & 65.60            & \multicolumn{1}{c|}{47.71}    & 68.95        & 61.35      \\
\multicolumn{1}{l|}{SMTA$_\text{{Transformer}}$}      & \textbf{66.55}   & \multicolumn{1}{c|}{\textbf{49.81}} & \textbf{70.05} & \textbf{61.73} \\ 
\bottomrule
\end{tabular}

}
% \vspace{-0.4cm}
% }
\end{table}

\noindent\textbf{Training.} 
The network is implemented using PyTorch \cite{paszke2019pytorch}. We use the Adam \cite{kingma2014adam} as the optimizer, with the initial learning rate as $5\times10^{-3}$ and the annealing ratio every 15 epochs as 0.1. The model is trained for 50 epochs and the mini-batch size is 16.

\subsection{Results and Comparisons}

\textbf{Methods to compare.} 
We compare two typical approaches with our proposed method.
All the methods will use the same multi-modal features presented above to assure fairness.
The main difference is in the first multi-modal fusion stage (the second part of Fig.\ref{tab:proposed_model}).
1) Normal LSTM. This method only uses the LSTM layer without attention mechanism as the first multi-modal fusion stage. 
The LSTM cells take as input the raw summed multi-modal features, without considering the modal-wise attention and the temporal context.
2) Existing attention methods. Those methods use attention modules (e.g., dot-product, non-local block, 
multi-head attention, transform encoder) to calculate attention weight by themselves in the first fusion stage, without temporal context. 
% to fuse the multi-modal features in the first fusion stage, without temporal context.
% 3) Synchronous Modal-Temporal Attention (SMTA). This is what we proposed in Sec.\ref{sec:approach}, in which the temporal context is used in the computation of modal attention matrix.

\noindent\textbf{Results.} 
% We evaluate the performance of the above methods on ERATO dataset on two subtasks, predicting the results with coarse and fine-grained categories. 
The performances in terms of Micro-F1 and Macro-F1 on ERATO for coarse and fine-grained categories are shown in Table~\ref{tab:performance}. 
As we could see, our proposed method SMTA with transformer encoder as 
attention module reaches the best result. 
Specifically, it achieves Micro-F1 of 66.55\%, Macro-F1 of 49.81\% 
for the task of 5 categories, and Micro-F1 of 70.05\%, Macro-F1 of 61.73\% for the task of 3 categories. 
Compared with LSTM, the result of SMTA with attention mechanism has a margin of at least 4\% on all metrics. 
The use of temporal context information also increases Macro-F1 by about 1\%, which is shown in Table~\ref{tab:performance}.

\begin{figure}[tbp]

\centering
\subfigure[the F1-score on task of 5 categories]{
\begin{minipage}[t]{0.5\linewidth}
\centering
\includegraphics[width=\linewidth]{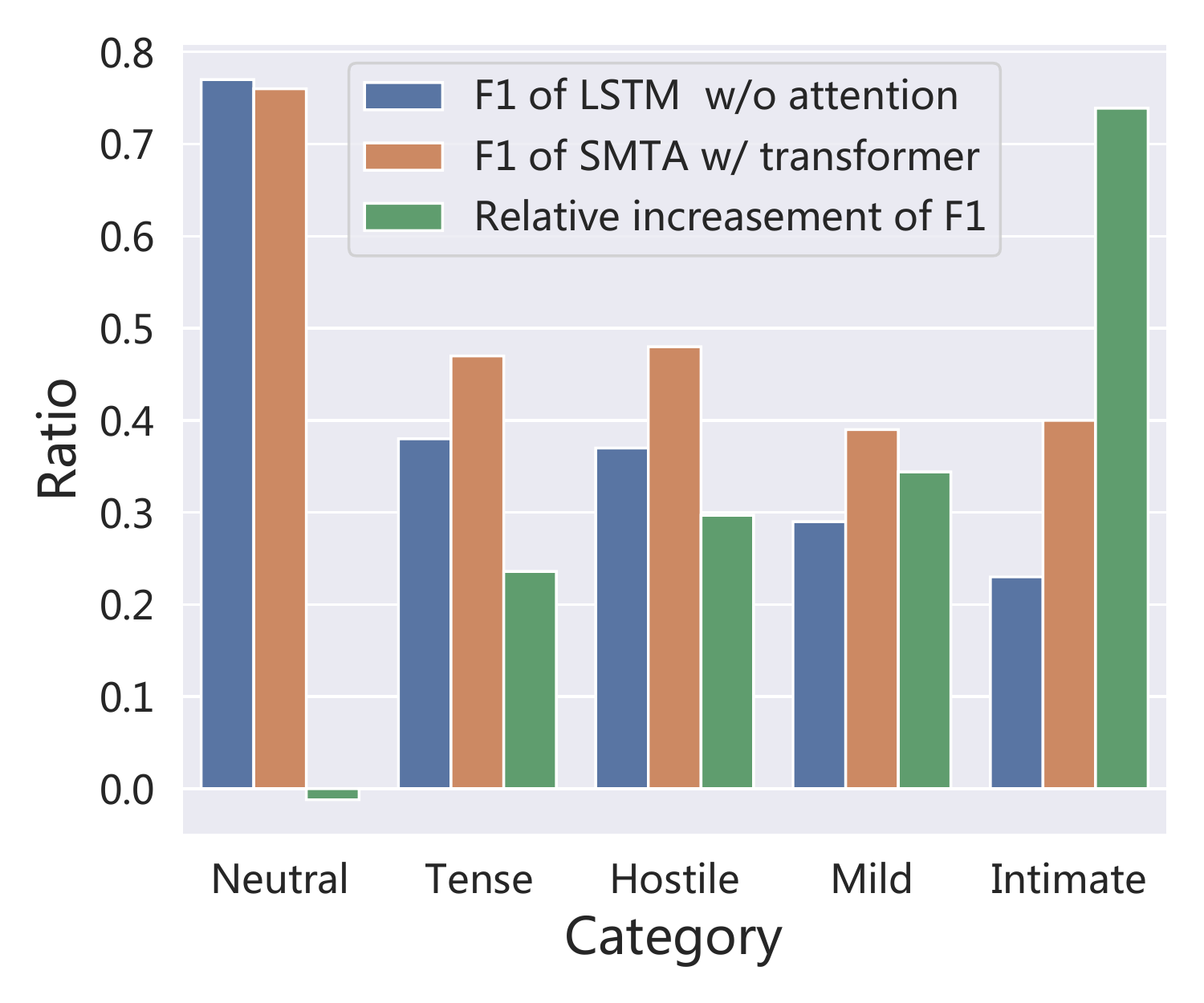}
% \caption{fig1}
\end{minipage}%
}%
\subfigure[the F1-score on task of 3 categories]{
\begin{minipage}[t]{0.5\linewidth}
\centering
\includegraphics[width=\linewidth]{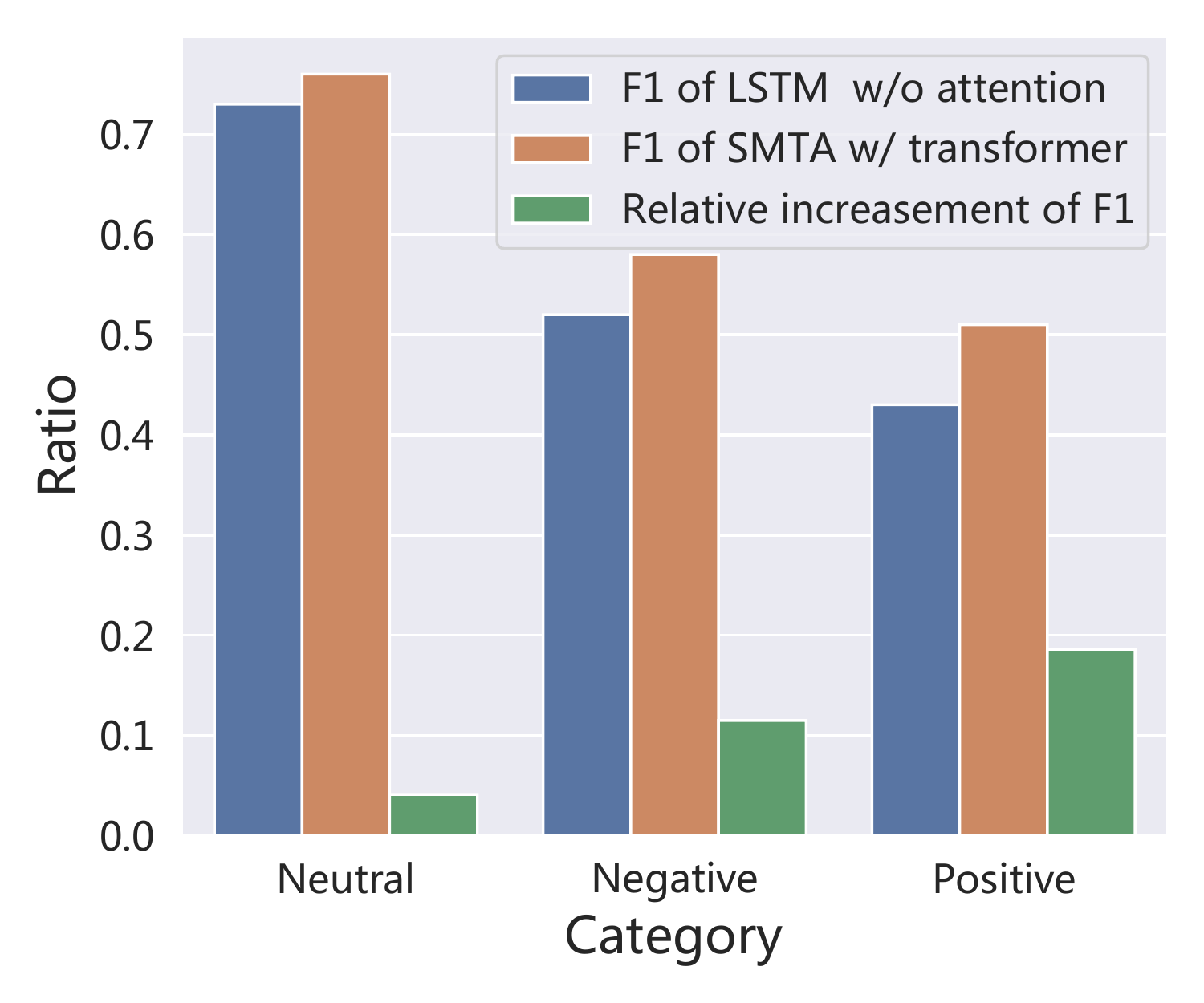}
%\caption{fig2}
\end{minipage}%
}%
\centering
% \vspace{-0.4cm}
\caption{Performance comparison between SMTA with transformer encoder and LSTM baseline.}
% \Description{}
\label{tab:Attention_stat}
% \vspace{-0.4cm}
\end{figure}

\begin{table*}[!t]
\caption{Evaluation of different modal features.}
\label{tab:Modal-performence}
\begin{tabular}{ccccccccc}
\toprule
\multirow{2}{*}{Index} & \multirow{2}{*}{Text} & \multirow{2}{*}{Audio} & \multicolumn{2}{c}{Visual} & \multicolumn{2}{c}{5 Category} &  \multicolumn{2}{c}{3 Category} \\
                       &                      &              & Background   & Person-wise                   & Micro-F1 & Macro-F1              & Micro-F1 & Macro-F1 \\ \midrule
\multicolumn{1}{c|}{1} & $\checkmark$         & $\checkmark$ & $\checkmark$ & \multicolumn{1}{c|}{}             & 63.80    & \multicolumn{1}{c|}{44.91} & 66.55         & 56.56          \\
\multicolumn{1}{c|}{2} & $\checkmark$         & $\checkmark$ &              & \multicolumn{1}{c|}{$\checkmark$} & 66.27    & \multicolumn{1}{c|}{48.93} & 69.48         & 60.52         \\
\multicolumn{1}{c|}{3} & $\checkmark$         &              & $\checkmark$ & \multicolumn{1}{c|}{$\checkmark$} & 64.94    & \multicolumn{1}{c|}{47.02} & 67.96         & 58.87         \\
\multicolumn{1}{c|}{4} & \multicolumn{1}{c}{} & $\checkmark$ & $\checkmark$ & \multicolumn{1}{c|}{$\checkmark$} & 65.19    & \multicolumn{1}{c|}{46.24} & 68.24         & 58.55         \\
\multicolumn{1}{c|}{5} &
  $\checkmark$ &
  $\checkmark$ &
  $\checkmark$ &
  \multicolumn{1}{c|}{$\checkmark$} &
  \multicolumn{1}{c}{\textbf{66.55}} &
  \multicolumn{1}{c|}{\textbf{49.81}} &
  \multicolumn{1}{c}{\textbf{70.05}} &
  \multicolumn{1}{c}{\textbf{61.73}} \\ \bottomrule
\end{tabular}
% }
\end{table*}

\begin{figure*}[t]
% \addtocounter{figure}{-1} %%%%%%%%%note1

\includegraphics[width=\linewidth]{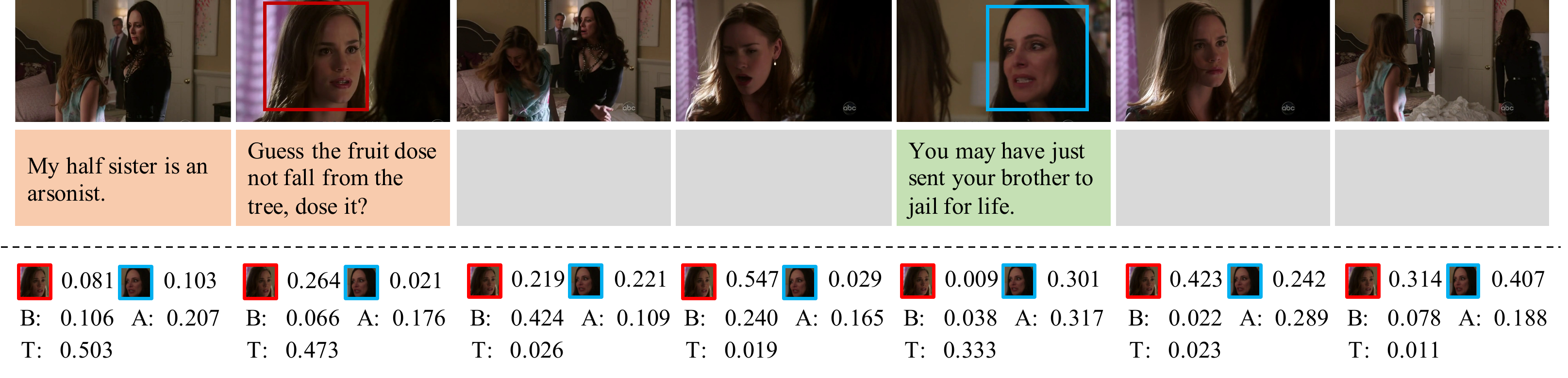}

\centering
% \vspace{-0.2cm}
\caption{Visualization of the attention weights for each modal feature, including the person-wise features, background (B), audio (A) and subtitle text (T). We could see that the intense dialogue 
makes the attention weight biased towards text and audio.
For the shot when the human behavior matters, the weights for characters and background features are increasing.}
% \vspace{-0.2cm}
\label{tab:Attention_example}
\end{figure*}

\begin{figure*}[!t]
% \addtocounter{figure}{-1} %%%%%%%%%note1
\includegraphics[width=\linewidth]{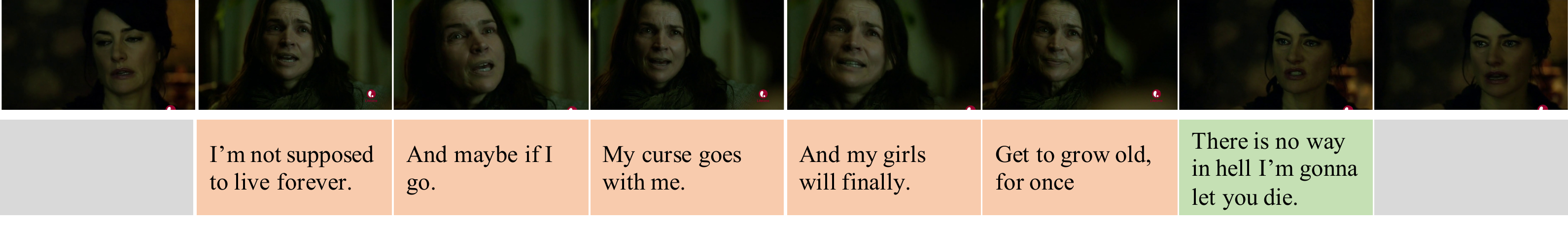}
\centering
% \vspace{-0.3cm}
\caption{Bad case example: Human can naturally infer the emotional relationship is positive from the dialogue content, but the model is difficult since the visual clue is opposite and the textual information is implicit.
}
% \vspace{-0.2cm}
\label{tab:reasoning}
\end{figure*}

\subsection{Analysis}
\textbf{How important are the paired interactive characters?} 
The goal of our work is to recognize the pairwise emotional relationship. 
Intuitively, the paired interactive characters should have a great impact on the experimental results. 
Comparing the results of the first and fifth row in Table~\ref{tab:Modal-performence}, 
the PERR performance decreases sharply by 5.17\% on Macro-F1 and 3.5\% on Micro-F1 for the 3 categories' subtask. 
Also, the same drop in metrics is found in the subtask of 5 categories, which indicates that 
paired interactive characters play an important role in PERR.

\textbf{Is multi-modality necessary for PERR task?} 
Multi-modal information is helpful for pairwise emotional relationship recognition, such as kissing as a visual cue 
and quarreling as an audio cue. Table ~\ref{tab:Modal-performence} shows that the best results can 
be achieved using different combinations of multi-modal features, including visual, audio and text. 
Comparing each row in Table~\ref{tab:Modal-performence} with the final result, we could roughly rank the importance 
of each modal feature by the performance gap on Macro-F1 in the 3 categories' subtask: the feature of person-wise (5.17\%), the textual feature (3.18\%), the audio feature (2.86\%) and the
background(1.21\%), which is intuitively consistent with the way of expressing affection in dramas or movies.

\textbf{Does the temporal information affect PERR?} As shown in Table~\ref{tab:performance}, 
no matter which attention mechanism is used, the performance of SMTA outperforms a 
margin of about 1\%. Especially in SMTA, combining temporal context and transformer encoder 
(last row in Table~\ref{tab:performance}), the Macro-F1 on 3 categories' 
subtask is 2.75\% higher than only transformer encoder. The main reason might be 
SMTA utilizes temporal context early in the modal-fusion stage, making the multi-modal information can be exchanged in the time dimension.

% \textbf{How does attention mechanism work on PERR?}
\textbf{What is the role of the attention mechanism in PERR?} The performance improvement of the attention mechanism 
for each category is represented in Fig.~\ref{tab:Attention_stat}. From Fig.~\ref{tab:Attention_stat}(a), 
the performance of all categories except neutral has been significantly improved compared with the baseline model without attention.
For harder categories with fewer samples such as hostile and intimate, the relative increasement for hostile is 29.7\% 
and for intimate is about 73.9\%.  
% In terms of hostile relativity increased by 29.7\%, while intimacy with 73.9\%. 
On the subtask with 3 categories, SMTA can acquire performance gain for each class: 4.1\% for neutral, 11.5\% 
for negative and 18.6\% for positive. 
Also, Fig.~\ref{tab:Attention_example} shows the attention weight for each modal in the SMTA unit.
We can see that when there are many dialogues, the attention weights for text and audio become large.
When the shot is focusing on the character's facial expression, the weight for the person-wise feature is larger.
This illustrates how attention works in fusing the different modal features. More examples are given in the appendix.
% \textbf{What can not be handled by current methods/modules?}

\textbf{Is SMTA applicable for other tasks?}
Our proposed SMTA model leverages the Modal-Temporal information simultaneously in multi-modal learning. Considering the PERR is a new task and cannot be fairly compared with existing methods,  
we report the experimental results of SMTA on the LIRIS-ACCEDE \cite{baveye2015liris} dataset, which is a multi-modality video dataset for predicting the audiences' emotional states in terms of Valence-Arousal metrics.
Our model achieves the best performance: MSE with 0.066, PCC with 0.454 on Valence prediction compared with the state-of-art.
Detailed results are in the appendix.

\textbf{Challenging Sample of PERR.}
There are many issues unsolved for PERR, such as large variation of character appearance, hidden but key clues investigation, etc. 
Fig. \ref{tab:reasoning} gives an example that the current model gives the wrong prediction.
In this clip, humans can easily infer that their emotional relationship is positive from the dialogue.
However, the visual features such as the facial expression, background music are sad or repressive. 
Only several key words in the dialogue give a strong indication of the real relationship.
It's hard for the current model to identify this case and we hope ERATO could help to develop more sophisticated methods to solve these issues.

\section{Conclusion}
In this paper, we propose a new task named Pairwise Emotional Relationship Recognition as well as the dataset called ERATO.
The task is to predict the emotional relationship between the two main characters in a video clip.
% This task is wildly spread and helpful for understanding the development of the story of dramas and movies.
%  develop a larger-scale multi-modal dataset,  
Built on top of dramas and movies, ERATO provides interaction-centric video clips with multi-shots, varied 
video length and multiple annotated modalities including video, audio and text. 
We also propose a baseline model with Synchronous Modal-Temporal Attention (SMTA) unit that considers the multi-modal fusion and temporal relation at the same time.
The comprehensive experiments of our proposed SMTA on ERATO validates the effectiveness of the method.
However, there is still a large room for PERR to improve, such as how to model the interactive sequences of the two characters explicitly, is matching the content of dialogue with the human expression helpful, is there a better way to handle the unbalanced modal information, etc.
We expect the PERR task can inspire more insights in both the field of affective computing and multi-modal learning.

%% The next two lines define the bibliography style to be used, and
%% the bibliography file.
\bibliographystyle{ACM-Reference-Format}
\balance
\bibliography{PERR}

\end{document}